%% file: main.tex
\documentclass[runningheads]{llncs}
\usepackage[T1]{fontenc}
\usepackage{graphicx}
\usepackage{booktabs}
\usepackage{amsmath}
\usepackage{amssymb}
\usepackage{mathtools}
\usepackage[rgb,dvipsnames]{xcolor}
\usepackage[hidelinks]{hyperref}
\usepackage{siunitx}
\usepackage{enumitem}
\usepackage[title]{appendix}

\makeatletter
\newcommand{\@chapapp}{\relax}%
\makeatother

\usepackage[caption=false]{subfig}

\usepackage[misc]{ifsym}
\newcommand{\corr}{(\Letter)}
\usepackage{mwe}

\usepackage{tikz,pgfplots}
\usetikzlibrary{fit,positioning,arrows,shapes,shapes.multipart,calc,arrows.meta,shapes.geometric,shapes.misc}
\usetikzlibrary{decorations.pathreplacing,angles,quotes,calligraphy}
\usetikzlibrary{automata,fit,positioning,arrows,shapes,shapes.multipart,calc,arrows.meta}
\usepgfplotslibrary{fillbetween}

\usepackage[toc,acronym,nonumberlist,nopostdot]{glossaries}
\makeindex
\makenoidxglossaries
\glsdisablehyper
\newacronym{DP}{DP}{differential privacy}
\newacronym{NDP}{NDP}{noiseless differential privacy}
\newacronym{LDP}{LDP}{local differential privacy}
\newacronym{PLRV}{PLRV}{privacy-loss random variable}
\newacronym{SPA}{SPA}{saddle-point accountant}
\newacronym{FL}{FL}{federated learning}

\newacronym{SecAgg}{SecAgg}{secure aggregation}
\newacronym{FP}{FP}{false positive}
\newacronym{FN}{FN}{false negative}
\newacronym{FPR}{FPR}{false positive rate}
\newacronym{FNR}{FNR}{false negative rate}

\newacronym{PDF}{PDF}{probability density function}
\newacronym{PMF}{PMF}{probability mass function}
\newacronym{CDF}{CDF}{cummulative distribution function}
\newacronym{CGF}{CGF}{cummulant generating function}

\newacronym{iid}{i.i.d.}{independent and identically distributed}

\newacronym{LLR}{LLR}{log-likelihood ratio}

\newacronym{LMIDP}{LMIDP}{local mutual-information differential privacy}

\newcommand{\FNR}{{\rm FNR}}
\newcommand{\FPR}{{\rm FPR}}

\newcommand{\revise}[1]{{#1}}

\input{math_commands_Hoang.tex}

\begin{document}

\title{Secure Aggregation is Not Private Against Membership Inference Attacks}

\titlerunning{Secure Aggregation is Not Private Against Membership Inference Attacks}

\author{Khac-Hoang Ngo\inst{1} \corr \and Johan \"Ostman\inst{2} \and Giuseppe Durisi\inst{1} \and \\ Alexandre Graell i Amat\inst{1}}

\authorrunning{K.-H. Ngo, J. \"Ostman, G. Durisi, and A. Graell i Amat}

\institute{Department of Electrical Engineering, Chalmers University of Technology, Gothenburg, Sweden  \email{\{ngok,durisi,alexandre.graell\}@chalmers.se} \and AI Sweden, Gothenburg, Sweden  \email{johan.ostman@ai.se}}

\tocauthor{Khac-Hoang~Ngo, Johan~\"Ostman, Giuseppe~Durisi, Alexandre~Graell~i~Amat}
\toctitle{Secure Aggregation is Not Private Against Membership Inference Attacks}	

\maketitle              

\begin{abstract}
Secure aggregation (SecAgg) is a commonly-used privacy-enhancing mechanism in federated learning, affording the server access only to the aggregate of  model updates while safeguarding the confidentiality of individual  updates.  Despite widespread claims regarding SecAgg's privacy-preserving capabilities, a formal analysis of its privacy is lacking, making such presumptions unjustified. In this paper, we delve into the privacy implications of SecAgg by treating it as a local differential privacy (LDP) mechanism for each local update. We design a simple attack wherein an adversarial server seeks to discern which update vector a client submitted, out of two possible ones, in a single training round of federated learning under SecAgg. By conducting privacy auditing, we assess the success probability of this attack and quantify the LDP guarantees provided by SecAgg. Our numerical results unveil that, contrary to prevailing claims, SecAgg offers weak privacy against membership inference attacks even in a single training round. Indeed, it is difficult to hide a local update by adding other independent local updates when the updates are of high dimension. Our findings underscore the imperative for additional privacy-enhancing mechanisms, such as noise injection, in federated learning.

\end{abstract}

\section{Introduction}  \label{sec:intro}

    \Gls{FL}~\cite{McM17} 
    allows multiple clients 
    to collaboratively train a machine learning model. In each training round, the clients share their local model updates with a central server, which then aggregates them to improve the global model. 
    Although raw data is not shared in the clear, vanilla \gls{FL} is prone to model-inversion attacks~\cite{Gei20} and membership-inference attacks~\cite{Nas19}. To mitigate such attacks, \gls{SecAgg}~\cite{Bon17} has been proposed, where the clients jointly mask their local model updates so that only the aggregate  is revealed to the server.  
    
    Many papers explicitly or implicitly assume that~\gls{SecAgg} provides strong privacy against honest-but-curious servers in a single round~\cite{Ull23,Hat23,Elk23b,So23}. 
    However, a formal analysis of the privacy offered by SecAgg is lacking, making this presumption unjustified. 
    \gls{SecAgg} has been combined with \gls{DP}~\cite{Dwo06_calibrating} to ensure that the server only sees the aggregate of the noisy local updates
    ~\cite{Aga18,Kai21}. 
    However, in these works, the privacy analysis does not account for SecAgg.
    It remains unclear how much privacy \gls{SecAgg} by itself provides for individual updates. 

\subsubsection{Main contributions.}
    \revise{
    We address the question: 
        \emph{how much privacy does \gls{SecAgg} by itself guarantee for the local updates?}
    Specifically, we formally analyze the privacy of SecAgg against membership inference attacks wherein the server aims to distinguish, from two potential update vectors, the one a client submitted in a single training round of FL with SecAgg. 
    Our approach consists in treating \gls{SecAgg} as a \gls{LDP} mechanism for each update, where the sum of the other clients' updates plays the role of a source of uncontrolled noise. We then characterize the privacy parameters $(\epsilon,\delta)$ for \gls{SecAgg} to satisfy $(\epsilon,\delta)$-\gls{LDP} via the following steps.
    \begin{itemize}[label=$\bullet$]
        \item We show that, under some practical assumptions, as the client population grows, the sum of the clients' updates converges to a Gaussian vector~(Theorem~\ref{th:noise_Gaussian}). We analyze the optimal privacy guarantee of the Gaussian mechanism with correlated noise~(Theorem~\ref{th:Gaussian_correlated_noise}). 

        \item We evaluate the optimal \gls{LDP} parameters of \gls{SecAgg} in some special cases (Theorem~\ref{th:dominating_pair} and Corollary~\ref{coro:tight_dominating_pair}) and verify that these parameters are close to that of a Gaussian mechanism, even for a small number of clients~(Fig.~\ref{fig:priv_curve}).

        \item Exploiting the similarity of \gls{SecAgg} and a Gaussian mechanism, we audit the privacy of \gls{SecAgg}. Specifically, we design a simple membership inference attack wherein the server regards \gls{SecAgg} as a Gaussian mechanism with correlated noise. We then evaluate the achievable \gls{FNR} and \gls{FPR} of this attack and use these values to compute a lower bound on the smallest $\epsilon$ for \gls{SecAgg} to satisfy $(\epsilon,\delta)$-\gls{LDP}.
    \end{itemize}
    We apply our privacy auditing procedure to federated averaging for a classification problem on the ADULT dataset~\cite{ADULT} and the EMNIST Digits dataset~\cite{Coh17EMNIST}. We show that both the \gls{FNR} and \gls{FPR} can be small simultaneously, and the audited $(\epsilon,\delta)$ are high.  Our results reveal that \gls{SecAgg} provides weak privacy even for a single training round. Indeed, it is difficult to hide a local update by adding other independent local updates when the updates are of high dimension. Therefore, \gls{SecAgg} cannot be used as a sole privacy-enhancing mechanism in \gls{FL}.
    }

\section{Related Work}
\subsubsection{Secure aggregation.}
    Based on cryptographic multi-party computation, SecAgg ensures that the central server sees only the aggregate of the clients' local updates, while individual updates are kept confidential. This is achieved by letting the clients jointly add randomly sampled masks to their updates via secret sharing, such that when the masked updates are aggregated, the masks cancel out~\cite{Bon17,Bel20}. 
    With \gls{SecAgg}, a client’s update is obfuscated by many other clients’ updates. However, the level of privacy provided by \gls{SecAgg} lacks a formal analysis. In~\cite{Elk23b}, this level was measured by the mutual information between a local update and the aggregated update. However, \revise{mutual information only measures the average privacy leakage and does not capture the threat to the most vulnerable data points. Furthermore, the bound provided in~\cite{Elk23b} is not explicit, i.e., not computable.
    }

\subsubsection{Differential privacy.}
    \gls{DP} is a rigorous privacy measure that quantifies the ability of an adversary to guess which dataset, out of two neighboring ones, a model was trained on~\cite{Dwo06_calibrating,Dwo14}. \gls{DP} is typically achieved by adding noise to the model/gradients obtained from the dataset~\cite{Aba16}. 
    A variant of \gls{DP} is \gls{LDP}~\cite{Kas11,Duc13}, where the noise is added to individual data points. When applied to achieve client-level privacy in 
    \gls{FL}, \gls{LDP} lets the clients add noise to their updates before sending the updates to the server. 

\subsubsection{Privacy attacks in \gls{FL} with \gls{SecAgg}.} 
    Model inversion attacks~\cite{Gei20} and membership-inference attacks~\cite{Nas19} have been shown to seriously jeopardize the integrity of FL. 
    When \gls{SecAgg} is employed, the server can perform disaggregation attacks to learn the individual data. A malicious server performs active attacks by suppressing the updates of non-target clients~\cite{Fow22,Boe23}. 
    For an honest-but-curious server, existing passive attacks require that the server leverages the aggregated model across many rounds~\cite{Lam21,Ker23}. 
    Differently from these works, we consider a \emph{passive} attack based only on the observation in a \emph{single round}. 

\section{Preliminaries} \label{sec:pre}

We denote random quantities with lowercase nonitalic letters, such as a scalar~$\rx$ 
and a vector~$\rvx$. The only exception is the \gls{PLRV}~$\rL$, which is in uppercase. 
Deterministic quantities are denoted with italic letters, such as a scalar~$x$ and a vector~$\vx$. 
We denote the multidimensional normal distribution with mean $\vmu$ and covariance matrix $\mSigma$ by $\sN(\vmu,\mSigma)$ and its \gls{PDF} evaluated at $\vx$ by $\sN(\vx; \vmu, \mSigma)$. We denote 
by $\Phi(x)$ the \gls{CDF} of the standard normal distribution $\sN(0,1)$, i.e., $\Phi(x) \triangleq \frac{1}{\sqrt{2\pi}} \int_{-\infty}^x e^{-u^2/2} \dif u$.
We denote by $[m:n]$ the set of integers from $m$ to $n$; 
$[n] \triangleq [1:n]$; 
 $(\cdot)^+ \triangleq \max\{0,\cdot\}$. Furthermore, $\ind{\cdot}$ denotes the indicator function, \revise{and $f(n) = o(g(n))$ means that $f(n)/g(n) \to 0$ as $n \to\infty$}.



Let $\phi$ be a decision rule of a hypothesis test between $\{H \colon\!$ the underlying distribution is $P\}$ and $\{H' \colon\!$ the underlying distribution is $Q\}$. Specifically, $\phi$ returns $0$ and $1$ in favor of $H$ and $H'$, respectively. 
A false positive (resp. false negative) 
occurs when $H$ (resp. $H'$) is true but rejected. 
The \gls{FPR} and \gls{FNR} of the test are given by $\alpha_\phi \triangleq \Exp[P]{\phi}$ and $\beta_\phi \triangleq 1 - \Exp[Q]{\phi}$, respectively.
\begin{definition}[Trade-off function] \label{def:trade_off}
     The trade-off function $T_{P,Q}(\cdot) \colon [0,1] \to [0,1]$ is the map from the \gls{FPR} to the corresponding minimum \gls{FNR} of the test between $P$ and $Q$, i.e., $T_{P,Q}(\alpha) \triangleq \inf_{\phi \colon \alpha_\phi \le \alpha} \beta_\phi$, $\alpha \in [0,1]$.
\end{definition}
We also write the trade-off function for the distributions of $\rx$ and $\ry$ as $T_{\rx,\ry}(\cdot)$.
We next state the definition of \gls{LDP}. 
\begin{definition}[{\gls{LDP}~\cite{Kas11,Duc13}}] \label{def:LDP}
    A mechanism $M$ satisfies $(\epsilon,\delta)$-\gls{LDP} if and only if, for every pair of data points $(\vx,\vx')$ and for every measurable set $\sE$, we have
    $
        \Prob{M(\vx) \in \sE} \le e^{\epsilon} \Prob{M(\vx') \in \sE} + \delta$. 
\end{definition}
For a mechanism $M$, we define the optimal \gls{LDP} curve $\delta_M(\eps)$ as the function that returns the smallest $\delta$ for which $M$ satisfies $(\eps,\delta)$-\gls{LDP}.
We next define a variant of \gls{LDP} that is built upon the trade-off function in a similar manner as $f$-\gls{DP}~\cite[Def.~3]{Don21}. 
\begin{definition}[{$f$-\gls{LDP}}] \label{def:f_LDP}
    $\!$For a 
    function $f$, a mechanism $M$ satisfies $\!f$-\gls{LDP} if~for every pair of data points $(\vx,\vx')$, we have that $T_{M(\vx), M(\vx')}(\alpha) \ge f(\alpha), \forall \alpha \in [0,1].$
\end{definition}
For a mechanism $M$, we define the optimal $f$-\gls{LDP} curve $f_M(\cdot)$ as the upper envelope of all functions $f$ such that $M$ satisfies $f$-\gls{LDP}. 
In the paper, we regard \gls{SecAgg} as a \gls{LDP} mechanism and provide bounds on both its optimal \gls{LDP} curve and optimal $f$-\gls{LDP} curve.

\section{Privacy Analysis of Secure Aggregation} \label{sec:priv_Agg}
We consider a FL scenario with $n+1$ clients and a central server. The model update of client $i$ can be represented as a vector $\rvx_i \in \bR^d$. Under \gls{SecAgg}, the server only learns the aggregate model update $\bar{\rvx}= \sum_{i=0}^n \rvx_i$,  while the individual updates $\{\rvx_i\}_{i=0}^n$ remain confidential. 

\revise{
\subsection{Threat Model}
\label{sec:threatmodel}
\subsubsection{Server.}
The server is honest and follows the \gls{FL} and \gls{SecAgg} protocols. We assume that it observes the exact sum $\bar{\rvx}$.
In practical \gls{SecAgg}, the clients discretize their updates (using, e.g., randomized rounding) and the server obtains a modulo sum. These operations introduce perturbations that can improve privacy~\cite{You23}. However, they do not capture the essence of \gls{SecAgg} which is to use the updates of other clients to obfuscate an individual update. The rounding and modulo operations can be applied even in a setting without \gls{SecAgg}. We ignore the perturbation caused by these operations to focus on the privacy obtainable by using the updates of other clients to obfuscate an update.

\subsubsection{Clients.} The clients are also honest. Client~$i$ computes the local model update $\rvx_i$ from the global model in the previous round and its local dataset. 
We also assume that each vector $\rvx_i$, $i\in [0:n]$, has correlated entries since these entries together describe a model, and that the vectors are mutually independent. 
The latter assumption holds in the first training round if the clients have independent local data. 
Furthermore, the independence assumption results in the best-case scenario for privacy as, if the vectors are dependent, the sum reveals more information about each vector. 
Therefore, the privacy level for the case of independent $\{\rvx_i\}_{i=0}^n$ acts as an upper bound on the privacy for the case of dependent vectors. So if privacy does not hold for independent data, it will also not hold when there is dependence.

\subsubsection{Privacy threat.}  
The server is curious. It seeks to infer the membership of a targeted client, say client~$0$, from the aggregate model updates $\bar{\rvx}$. We consider a membership inference game~\cite{Ye22} where: i) a challenger selects a pair of possible local updates $(\vx_0,\vx_0')$ of client~$0$, one of which is used in the aggregation, and sends this pair to the server, ii) the server observes $\bar{\rvx}$ and guesses if $\vx_0$ or $\vx_0'$ was submitted by client~$0$. Note that this attack can be an entry point for the server to further infer the data of client~$0$.
Our goal is to quantify the capability of \gls{SecAgg} in mitigating this attack.

\subsection{\gls{SecAgg} as a Noiseless \gls{LDP} Mechanism}
Hereafter, we focus on client~$0$; the analysis for other clients follows similarly. 
Our key idea is to view \gls{SecAgg} through the lens of noiseless \gls{DP}~\cite{Bha11}, where the contribution of other clients can be seen as noise and no external (controlled) noise is added. More precisely, for client $0$, \gls{SecAgg} plays the role of the mechanism
\begin{equation}
    M(\rvx_0) = \rvx_0 + \rvy, \label{eq:mechanism_client_0}
\end{equation}
where $\rvy = \sum_{i=1}^n\rvx_i$  is a source of uncontrolled noise.

The aforementioned membership inference game can be cast as follows: given $M(\rvx_0)$, the server guesses whether it came from $P_{M(\rvx_0) | \rvx_0 = \vx_0}$ or $P_{M(\rvx_0) | \rvx_0 = \vx_0'}$ for the worst-case pair $(\vx_0,\vx_0')$. This game is closely related to the \gls{LDP} framework. First, the tradeoff between the \gls{FPR} and \gls{FNR} of the server's guesses is captured by the $f$-\gls{LDP} guarantee of $M$. Second, as $M$ achieves a stronger $(\epsilon,\delta)$-\gls{LDP} guarantee, the distributions $P_{M(\rvx_0) | \rvx_0 = \vx_0}$ and $P_{M(\rvx_0) | \rvx_0 = \vx_0'}$ become more similar, and the hypothesis test between them becomes harder. 
Therefore, we shall address the following question: how much \gls{LDP} or $f$-\gls{LDP} does \gls{SecAgg} guarantee for client~$0$? 
Specifically, we shall establish bounds on the optimal \gls{LDP} curve $\delta_M(\epsilon)$ and optimal $f$-\gls{LDP} curve $f_M(\cdot)$ of the mechanism $M$.
}

\subsection{Asymptotic Privacy Guarantee} \label{sec:asymptotic}
Let us first focus on the large-$n$ regime. The following asymptotic analysis will be used as inspiration \revise{for our privacy auditing procedure} to establish a lower bound on the \gls{LDP} curve in Section~\ref{sec:priv_auditing}. 

We assume that the $\ell_2$ norm of the vectors $\{\rvx_i\}_{i=1}^n$ scales as $o(\sqrt{n})$, which holds if, e.g., $d$ is fixed. 
In this case, $\rvy$ converges to a Gaussian vector when $n\to\infty$, as stated in the next theorem.
\begin{theorem}[Asymptotic noise distribution] \label{th:noise_Gaussian}
    Assume that $\{\rvx_i\}_{i=1}^n$ are independent, $\|\rvx_i\|_2 = o(\sqrt{n})$ for $i \in [n]$, and $\frac{1}{n}\sum_{i=1}^n\Cov[\rvx_i] \to \mSigma$ as $n \to \infty$. Then $\frac{1}{\sqrt{n}}\big(\!\sum_{i=1}^n \!\rvx_i - \Exp{\sum_{i=1}^n \!\rvx_i}\!\big)$ converges in distribution to $\sN(\mathbf{0},\mSigma)$ as $n\to\infty$.
\end{theorem}
\begin{proof}
    Theorem~\ref{th:noise_Gaussian} follows by applying the multivariate Lindeberg-Feller central limit theorem~\cite[Prop.~2.27]{Van00} to the triangular array $\big\{\frac{\rvx_i}{\sqrt{n}}\big\}_{n,i}$, upon verifying the Lindeberg condition 
        $\lim\limits_{n\to\infty} \sum_{i=1}^n \Exp{\frac{\|\rvx_i\|_2^2}{n} \ind{\frac{\|\rvx_i\|_2}{\sqrt{n}} > \varepsilon}} = 0, \forall \varepsilon > 0$.
    Since $\|\rvx_i\|_2 = o(\sqrt{n})$, i.e., $\|\rvx_i\|_2/\sqrt{n} \to 0$ as $n\to\infty$, 
    this condition indeed holds. 
    \qed
\end{proof}

Theorem~\ref{th:noise_Gaussian} implies that, when $n$ is large, under the presented assumptions, the mechanism $\widetilde M(\vx_0) = M(\vx_0) - \Exp{\rvy}$ behaves like a Gaussian mechanism with noise distribution $\sN(\mathbf{0},\mSigma_\rvy)$, where $\mSigma_\rvy$ is the covariance matrix of $\rvy$. Furthermore, 
since the map from $M$ to $\widetilde M$ is simply a shift by a fixed vector $\Exp{\rvy}$, i.e., it is a bijection, we have from the post-processing property\footnote{If a mechanism $M$ satisfies $(\epsilon,\delta)$-\gls{LDP}, then so does $h \circ M$ for a mapping $h$ that is independent of $M$. The proof of this result is similar to the proof of the post-processing property of \gls{DP}~\cite[Prop.~2.1]{Dwo14}.} that the optimal \gls{LDP} curve of $M$ is the same as that of $\widetilde M$. 

We now provide privacy guarantees for a Gaussian mechanism with correlated noise, to capture the correlation between the entries of the vectors $\rvx_i$. The next theorem, proved in Appendix~\ref{proof:Gaussian_correlated_noise}, is an extension of the optimal privacy curve of the uncorrelated Gaussian mechanism~\cite[Theorem~8]{Bal18}. 
%
\begin{theorem}[Correlated Gaussian mechanism] \label{th:Gaussian_correlated_noise}
    Consider the mechanism $G(\rvx) = \rvx + \rvy$ where $\rvx$ belongs to a set $\sS_d \subset \bR^d$, and $\rvy \sim \sN(\mathbf{0}, \mSigma_\rvy)$. The optimal \gls{LDP} curve of $G$ is 
    \begin{equation}
        \delta_G(\epsilon) = \Phi\Big(\frac{\Delta}{2} - \frac{\epsilon}{\Delta}\Big) - e^{\epsilon}\Phi\Big(-\frac{\Delta}{2} - \frac{\epsilon}{\Delta}\Big)\label{eq:priv_curve_Gaussian_correlated_noise}
    \end{equation}
    where $\Delta = \max_{\vx,\vx' \in \sS_d} \Delta_{\vx,
    \vx'}$ with 
    \begin{equation}
        \Delta_{\vx,\vx'} \triangleq \sqrt{(\vx-\vx')^\T\mSigma_\rvy^{-1}(\vx - \vx')}. \label{eq:worst_case_pair}
    \end{equation}
\end{theorem}
In Section~\ref{sec:upper_bound}, we shall verify the similarity between the privacy of \gls{SecAgg} and that of the Gaussian mechanism $G$ 
via numerical examples. 

\revise{Parameter $\Delta$ is the maximum Euclidean distance between a pair of input vectors transformed by  matrix $\mSigma^{-1/2}$ (similar to the whitening transformation). It plays the same role as the ratio of the sensitivity and the noise standard deviation in the case of uncorrelated noise~\cite{Bal18}.} We remark that the privacy guarantee of $G$ is weakened as~$\Delta$ increases: for a given $\epsilon$, $\delta_G(\epsilon)$ increases with~$\Delta$. To achieve small $\epsilon$ and $\delta$, we need $\Delta$ to be small. 
%
The impact of $\Delta$ can also be seen via \revise{the hypothesis test associated to the considered membership inference game}. Consider an adversary that observes an output $\vz$ of $G$ and tests between $\{H: \text{$\vz$ came from $P_{G(\rvx) | \rvx = \vx}$}\}$ and $\{H': \text{$\vz$ came from $P_{G(\rvx) | \rvx = \vx'}$}\}$. This is effectively a test between $\sN(\vx, \mSigma_\rvy)$ and $\sN(\vx', \mSigma_\rvy)$. The trade-off function for this test is stated in the following proposition, which is proved in Appendix~\ref{proof:tradeoff_Gaussian}.
\begin{proposition} \label{prop:tradeoff_Gaussian}
        $T_{\sN(\vx,\mSigma_\rvy), \sN(\vx',\mSigma_\rvy)}(\alpha)  = \Phi\big(\Phi^{-1}(1-\alpha) - \Delta_{\vx,\vx'}\big)$, $\alpha \in [0,1]$.
\end{proposition}
The trade-off function decreases with $\Delta_{\vx,\vx'}$. A large $\Delta_{\vx,\vx'}$ facilitates the distinguishability of the pair $(\vx,\vx')$, and thus weakens the privacy guarantee. Furthermore, the worst-case pair $(\vx,\vx')$ that minimizes the trade-off function 
is given by the maximizer of $\Delta_{\vx,\vx'}$. It follows that the optimal $f$-\gls{LDP} curve of the Gaussian mechanism $G$ is $f_G(\alpha) = \Phi\big(\Phi^{-1}(1-\alpha) - \Delta\big)$, $\alpha \in [0,1]$.

\subsection{Upper Bounding $\delta_M(\epsilon)$ via Dominating Pairs of Distributions} \label{sec:upper_bound}
We now upper-bound the optimal \gls{LDP} curve of~$M$ in~\eqref{eq:mechanism_client_0} for finite~$n$. \revise{We shall then consider the case in which the upper bound is tight and verify the convergence of \gls{SecAgg} to a Gaussian mechanism.}

%
We define the hockey-stick divergence with parameter $\alpha$ between two probability measures $P$ and $Q$ as
        $\sfE_\alpha(P\|Q) = 
         \sup_{\sE} (P(\sE) - \alpha Q(\sE))$. 
We also write the hockey-stick divergence between the distributions of $\rx$ and $\ry$ as $\sfE_\alpha(\rx \|\ry)$. The condition for \gls{LDP} in Definition~\ref{def:LDP} is equivalent to $\textstyle\sup_{\vx \ne \vx'} \sfE_{e^\eps}(M(\vx) \| M(\vx')) \le \delta.$
Therefore, the optimal \gls{LDP} curve of mechanism $M$ is given by 
\begin{equation}
\delta_M(\eps) = \textstyle\sup_{\vx \ne \vx'} \sfE_{e^\eps}(M(\vx) \| M(\vx'))\,.
\end{equation}

A pair of measures $(P,Q)$ is called a dominating pair of distributions for $M$~if
 \begin{equation} \label{eq:dominating_pair}
    \textstyle\sup_{\vx \ne \vx'} \sfE_{e^\eps}(M(\vx) \| M(\vx')) \le \sfE_{e^\eps}(P \| Q),\quad \forall \epsilon \ge 0\,.
 \end{equation}
If equality is achieved in~\eqref{eq:dominating_pair} 
for every $\epsilon \ge 0$, then $(P,Q)$ is said to be a tightly dominating pair of distributions for $M$.
For each dominating pair $(P,Q)$, we associate a privacy-loss random variable $\rL \triangleq \ln \frac{\dif P}{\dif Q}(\rvy)$ with $\rvy \sim P$, 
where $\frac{\dif P}{\dif Q}$ is the Radon-Nikodym derivative. 
We have that 
\begin{equation}
    \sfE_{e^\epsilon}(P \| Q) = \Exp{(1-e^{\epsilon - \rL})^+} \triangleq \delta_\rL(\epsilon)\,.
\end{equation}
It follows readily that 
$\delta_\rL(\epsilon)$ is an upper bound on the optimal \gls{LDP} curve $\delta_M(\epsilon)$. 


Without a known distribution of $\rvy$, it is challenging to characterize a dominating pair of distributions for the mechanism $M$ in~\eqref{eq:mechanism_client_0}. In the next theorem, proved in~\ref{app:dominating_pair}, we make some assumptions on $P_\rvy$ to enable such characterization. 
\begin{theorem}[Dominating pair of distributions]\label{th:dominating_pair}
    Let $\rvx_0 = (\rx_{01}, \rx_{02}, \dots, \rx_{0d})$ and assume that $\underline r_j \le \rx_{0j} \le \overline r_j$, $j\in [d]$. Assume further that $\rvy$ has independent entries, i.e., $P_\rvy = P_{\ry_1} \times \dots \times P_{\ry_d}$, and that the marginal probabilities $\{P_{\ry_j}\}$ are log-concave and symmetric.\footnote{$P_{\ry_j}$ is symmetric if there exists a $y^*$ such that $P_{\ry_j}(\sA + y^*) = P_{\ry_j}(-\sA + y^*)$ for every subset $\sA$ of the support of $\ry_j$. Here, $-\sA \triangleq \{-y \colon y \in \sA\}$.} Then, a dominating pair of distributions for the mechanism $M(\rvx_0)$ in~\eqref{eq:mechanism_client_0} is given by $(P_{\underline r_1 + \ry_1} \times \dots \times P_{\underline r_d + \ry_d}, P_{\overline r_1 + \ry_1} \times \dots \times P_{\overline r_d + \ry_d})$. 
\end{theorem}
The family of log-concave distributions includes the typical noise distributions in \gls{DP}, namely, the Gaussian and Laplace distributions, as well as many other common distributions, e.g., the exponential and uniform distributions~\cite{Bag06}. If each vector $\rvx_i, i\in [n],$ has independent entries following a log-concave distribution, then so does the sum $\rvy = \sum_{i=1}^n \rvx_i$, because log-concavity is closed under convolutions~\cite{Mer98}. 
Under the presented assumptions, Theorem~\ref{th:dominating_pair} allows us to characterize an upper bound on the \gls{LDP} curve of ${M}$  as
$
    \delta_\rL(\epsilon)
$
with $\rL = \sum_{j=1}^d 
\ln \frac{P_{\underline r_j + \ry_j}(\rz_j)}{P_{\overline r_j + \ry_j}(\rz_j)}$ where $\rz_j \sim P_{\underline r_j + \ry_j}$, $j\in [d]$. 


\begin{corollary} \label{coro:tight_dominating_pair}
    If the support of $\rvx_0$ contains $(\underline r_1,\dots,\underline r_d)$ and $(\overline r_1,\dots,\overline r_d)$, the dominating pair of distributions in Theorem~\ref{th:dominating_pair} becomes tight, and the resulting upper bound $\delta_\rL(\epsilon)$ is the optimal \gls{LDP} curve.
\end{corollary}

\begin{figure}[t]
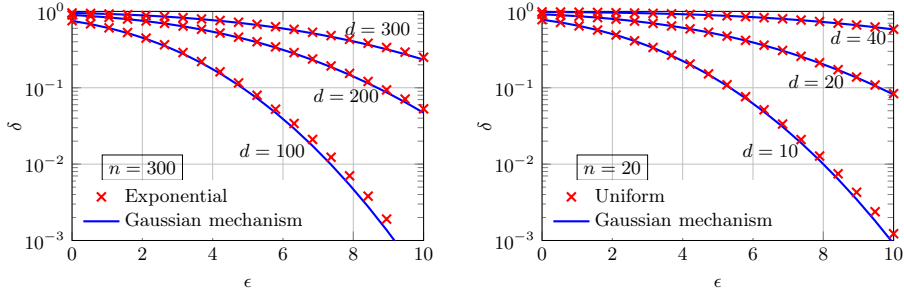

    \centering
	\include{fig/priv_curve_unif_n20}
    \vspace{-.5cm}
    \caption{The optimal \gls{LDP} curve of $M$ in~\eqref{eq:mechanism_client_0} where each $\rvx_i$ has independent entries, compared with the Gaussian mechanism $G$ with the same noise covariance matrix. 
    }
    \label{fig:priv_curve}
    \vspace{-.3cm}
\end{figure}

We now use Corollary~\ref{coro:tight_dominating_pair} to evaluate the optimal \gls{LDP} curve of mechanism $M$ in~\eqref{eq:mechanism_client_0} when each $\rvx_i$ has independent entries. We aim to \revise{verify the convergence of \gls{SecAgg} to a Gaussian mechanism implied by Theorem~\ref{th:noise_Gaussian} and} understand how the \gls{LDP} curve depends on the model size~$d$. We consider two cases. In the first case, the entries follow the exponential distribution with parameter $1$, and thus $\rvy$ has independent entries following the Gamma distribution with shape $n$ and scale $1$. For convenience, we further assume that $\rvx_{0}$ is truncated such that $0\le \rx_{0j} \le 4$, $j\in [d]$. In the second case, the entries are uniformly distributed in $[-1/2,1/2]$, and thus $\rvy$ has independent entries following the shifted Irwin–Hall distribution with \gls{PDF}
$p_{\ry_i}(y) = \frac{1}{(n-1)!} \sum_{k=0}^n (-1)^k \binom{n}{k}\big[(y+n/2-k)^+\big]^{n-1}$. Both cases satisfy the conditions of Corollary~\ref{coro:tight_dominating_pair}. We can therefore obtain the optimal \gls{LDP} curves and depict them in Fig.~\ref{fig:priv_curve}. We also show the optimal \gls{LDP} curve of the Gaussian mechanism $G$ with the same noise covariance matrix $\mSigma_\rvy$. We see that the optimal \gls{LDP} curve of $M$ is indeed close to that of $G$, even for a small value of $n$ in the second case. Furthermore, although Theorem~\ref{th:noise_Gaussian} assumes a fixed $d$ and $n \to \infty$, Fig.~\ref{fig:priv_curve} suggests that $M$ behaves similarly to a Gaussian mechanism even for large~$d$. Remarkably, for a given $\epsilon$, the parameter~$\delta$ increases rapidly with~$d$, indicating that the privacy of \gls{SecAgg} is weak for high-dimensional models. 

\subsection{Lower Bounding $\delta_M(\epsilon)$ and Upper Bounding $f_M(\cdot)$ 
via Privacy Auditing} \label{sec:priv_auditing}
In practical \gls{FL}, the updates typically have a distribution that does not satisfy the conditions of Theorem~\ref{th:dominating_pair} and is not known in closed form. Therefore, we now establish a numerical routine to compute a lower bound on the optimal \gls{LDP} curve and an upper bound on the optimal $f$-\gls{LDP} curve of $M$. \revise{The proposed numerical routine exploits the similarity of SecAgg and a Gaussian mechanism as discussed in Sections~\ref{sec:asymptotic} and~\ref{sec:upper_bound}.  The}  bounds are based on the following result. 

\begin{proposition}[\gls{LDP} via the trade-off function] \label{prop:LDP_tradeoff}
    A mechanism $M$ satisfies $(\epsilon,\delta)$-\gls{LDP} if and only if for every $\alpha \in [0,1]$,
    \begin{equation}
        \epsilon \ge \ln \max\bigg\{\frac{1-\delta-\alpha}{\inf_{\vx \ne \vx'} T_{M(\vx),M(\vx')}(\alpha)}, \frac{1-\delta-\inf_{\vx \ne \vx'} T_{M(\vx),M(\vx')}(\alpha)}{\alpha}\bigg\}\,.
        \label{eq:LDP_tradeoff}
    \end{equation}
\end{proposition}
The proof of Proposition~\ref{prop:LDP_tradeoff} follows from similar arguments for \gls{DP} in~\cite[Thm.~2.1]{Kai17}. This proposition implies that, if a pair $({\rm FPR},{\rm FNR})$ is achievable for some decision rule $\phi$ between the distributions of $M(\vx)$ and $M(\vx')$ for some $(\vx,\vx')$, then the mechanism does not satisfy $(\epsilon,\delta)$-\gls{LDP} for 
$\delta \in [0,1]$ and 
\begin{equation}
    \epsilon < \ln \max\bigg\{\frac{1-\delta-{\rm FPR}}{\rm FNR}, \frac{1-\delta-{\rm FNR}}{\rm FPR}\bigg\}\,. \label{eq:priv_auditing}
\end{equation}
This gives a lower bound on the optimal \gls{LDP} curve of $M$. 
Furthermore, it follows readily from Definition~\ref{def:f_LDP} that,  for an achievable pair $({\rm FPR},{\rm FNR})$ of the mentioned test, the mechanism does not satisfy $f$-\gls{LDP} for any trade-off function $f$ such that $f(\FPR) < \FNR$. Therefore, a collection of achievable pairs $(\FPR, \FNR)$ constitutes an upper bound on the optimal $f$-\gls{LDP} curve of $M$. 

We shall use this result to perform {\it privacy auditing}~\cite{Jag20,Ye22}. Specifically, \revise{following the defined membership inference game (see privacy threat in Section~\ref{sec:threatmodel}), we conduct a hypothesis test} between $\{H \colon \textnormal{$\vz$ came from $P_{M(\rvx_0) | \rvx_0 = \vx_0}$}\}$ and $\{H' \colon \textnormal{$\vz$ came from $P_{M(\rvx_0) | \rvx_0 = \vx_0'}$}\}$ for a given output $\vz$ of $M$. That is, we select a pair $(\vx_0,\vx_0')$ \revise{for the challenger} and a decision rule~$\phi$ \revise{for the server}. We then evaluate the achievable pair $(\FPR, \FNR)$, and obtain therefrom a lower bound on the optimal \gls{LDP} curve and an upper bound on the optimal $f$-\gls{LDP} curve of $M$. 
To design the attack, we draw inspiration from the asymptotic analysis in Section~\ref{sec:asymptotic} as follows. 

We consider the likelihood-ratio test, i.e., the test rejects $H$ if 
\begin{equation}
    \ln \frac{P_{M(\rvx_0) \given \rvx_0}(\vz \given \vx_0)}{P_{M(\rvx_0) \given \rvx_0}(\vz \given \vx_0')} = \ln \frac{P_{\rvy}(\vz - \vx_0)}{P_{\rvy}(\vz - \vx_0')} \le \theta
\end{equation}    
for a given threshold $\theta$.  
We choose the input pair $(\vx_0,\vx_0')$ as the worst-case pair, i.e., $(\vx_0,\vx_0') = \argmin_{\vx,\vx'} T_{M(\vx),M(\vx')}(\alpha), ~\forall \alpha \in [0,1]$. 
However, the trade-off function is not known in closed-form in general, and thus finding the worst-case pair is challenging. 
Motivated by Theorem~\ref{th:noise_Gaussian}, we treat $\rvy$ as a Gaussian vector with the same mean $\vmu_{\rvy}$ and covariance $\mSigma_\rvy$. 
We thus approximate the trade-off function $T_{M(\vx),M(\vx')}(\cdot)$ by $T_{\sN(\vx,\mSigma_\rvy),\sN(\vx',\mSigma_\rvy)}(\cdot)$, and choose $(\vx_0,\vx_0')$ as the minimizer of the latter. Using  Proposition~\ref{prop:tradeoff_Gaussian}, we have that
\begin{equation} \label{eq:input_pair_auditing}
    (\vx_0,\vx_0') = \argmax_{\vx,\vx' \in \sX_0} \Delta_{\vx,
    \vx'}
\end{equation}
where $\sX_0$ is the support of $\rvx_0$.

If the \revise{server} does not know $P_\rvy(\vy)$ in closed form, we let it approximate $P_\rvy(\vy)$ as $\sN(\vy;  \vmu_\rvy, \mSigma_\rvy)$. That is, the test rejects $\vx_0$ if 
\begin{equation} \label{eq:test_metric_normal_approx}
    \ln \frac{\sN(\vz - \vx_0; \vmu_\rvy, \mSigma_\rvy)}{\sN(\vz - \vx_0'; \vmu_\rvy, \mSigma_\rvy)} \le \theta\,.
\end{equation}
Moreover, if the \revise{server} does not know $\vmu_\rvy$ and $\mSigma_\rvy$ but can generate samples from $P_\rvy$, we let it estimate $\vmu_\rvy$ and $\mSigma_\rvy$ as the sample mean and sample covariance matrix, and use these estimates instead of the true values in~\eqref{eq:input_pair_auditing} and~\eqref{eq:test_metric_normal_approx}.



We evaluate the $\FNR$ and $\FPR$ of the test via Monte-Carlo simulation. Specifically, we repeat the test $N\sub{s}$ times and count the number of false negatives $N\sub{FN}$ and the number false positives $N\sub{FP}$. We obtain a high-confidence upper bound on \gls{FNR} using the Clopper-Pearson method~\cite{Clop34} as
$\overline{\rm FNR} = B(1-\gamma/2; N\sub{FN} + 1, N\sub{s} - N\sub{FN})$, where $B(x; a,b)$ is the quantile of the Beta distribution with shapes $(a,b)$, and $1-\gamma$ is the confidence level. A high-confidence upper bound $\overline{\rm FPR}$ on \gls{FPR} is obtained similarly. By varying the threshold $\theta$, we obtain an empirical trade-off curve $\overline{\rm FNR}$ vs. $\overline{\rm FPR}$. This curve is an upper bound on the optimal $f$-\gls{LDP} curve of \gls{SecAgg}. For a given $\delta \in [0,1]$, we also compute a lower confidence bound on $\epsilon$ for \gls{SecAgg} to satisfies $(\epsilon,\delta)$-\gls{LDP}. Specifically, we use $\overline{\rm FNR}$ and $\overline{\rm FPR}$ in place of ${\rm FNR}$ and $\rm FPR$ in~\eqref{eq:priv_auditing}. Note that $\overline{\rm FNR}$ and $\overline{\rm FPR}$ are lower bounded by $B(1-\gamma/2; 1, N\sub{s})$ even if $N\sub{FN} = N\sub{FP} = 0$. Therefore, the estimated $\epsilon$ is upper bounded by 
    $\overline \epsilon_\delta \triangleq \ln \frac{1 - \delta - B(1-\gamma/2; 1, N\sub{s})}{B(1-\gamma/2; 1, N\sub{s})}.$  
That is, it is impossible to audit an arbitrarily large $\epsilon$ with a finite number of trials.

\section{Experiments and Discussion} \label{sec:experiment}

\subsubsection{Experimental setting.} We consider 
federated averaging with $n_{\rm tot} = 100$ clients out of which $n+1$ clients are randomly selected in each round. The experiments\footnote{\revise{The code is available at \href{https://github.com/khachoang1412/SecAgg_not_private}{https://github.com/khachoang1412/SecAgg\_not\_private}.}}
are conducted for a classification problem on the ADULT dataset~\cite{ADULT} and the EMNIST Digits dataset~\cite{Coh17EMNIST}. The ADULT dataset contains $\num{30162}$ entries with $104$ features; the entries belong to two classes with $7508$ positive labels and $\num{22654}$ negative labels. The EMNIST Digits dataset contains $\num{280000}$ images of size $28\times 28$ of handwritten digits belonging to $10$ balanced classes. \revise{We allocate the training samples between $n_{\rm tot}$ clients according to a latent Dirichlet allocation model 
with concentration parameter $\omega$. Here, with $\omega \to \infty$, the training samples are distributed evenly and uniformly between the clients; with $\omega \to 0$, each client holds samples from only one class.} We consider a single-layer neural network and use the cross-entropy loss and stochastic gradient descent with a learning rate of $0.01$ and batch size of $64$. 
The model size is $d = 210$ for ADULT and $d = 7850$ for EMNIST Digits. 

We focus on the first round, containing one local epoch, of federated averaging and perform privacy auditing for a fixed initial model, which is known to the server. Note that performing an attack in the first round is the most challenging because, in later rounds, the server accumulates more observations. Let $\{\rvx_i\}_{i=0}^n$ be the local updates of the selected clients in the first round. 
The server does not know the distribution of $\{\rvx_i\}_{i=0}^n$ in closed form, but can sample from this distribution by simulating the learning scenario. Note that it is a common assumption in membership inference attacks that the adversary can sample from the population~\cite{Ye22}. We let the server compute the sample mean $\hat \vmu_\rvx$ and sample covariance matrix $\widehat \mSigma_\rvx$ from $\num{25000}$ samples of $\rvx_i$, then estimate the mean and covariance matrix of $\rvy = \sum_{i=1}^n \rvx_i$ as $\hat \vmu_\rvy = n\hat\vmu_\rvx$ and $\widehat \mSigma_\rvy = n \widehat \mSigma_\rvx$. The server then uses $\hat \vmu_\rvy$ and $\widehat \mSigma_\rvy$ for privacy auditing, as described in Section~\ref{sec:priv_auditing}. Following~\eqref{eq:input_pair_auditing}, we find the worst-case input pair $(\vx_0,\vx_0')$ by searching for the maximizer of $\Delta_{\vx_0,\vx_0'}$ among $5000$ and $1000$ samples of $\rvx_0$ for the ADULT and EMNIST Digits datasets, respectively. The Clopper-Pearson confidence level is $1-\gamma = 95\%$. For a given initial model, we consider $N_{\rm s} = 5000$ trials with random data partition and batch selection. In the simulation results, we report the average of $\overline{\FPR}$, $\overline{\FNR}$, and the audited $(\epsilon,\delta)$ over $10$ and $5$ initial models for the ADULT and EMNIST Digits datasets, respectively.
\vspace{1ex}

\subsubsection{\revise{Homogeneous data partitioning.}} 
\begin{figure}[t!]
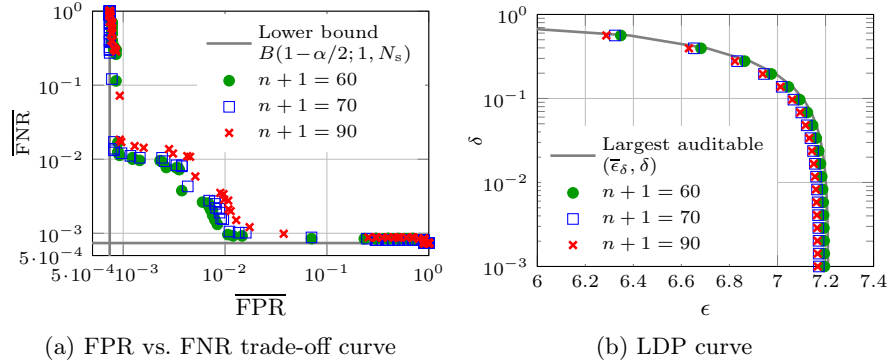

    \centering
    \include{fig/fig_ADULT}
    \vspace{-4ex}
    \caption{The audited \gls{FPR} vs. \gls{FNR} trade-off and \gls{LDP} curve, averaged over $10$ initial models, for \gls{SecAgg} in federated learning on the ADULT dataset \revise{with homogeneous data partitioning}. Here, $d = 210$.}
    \label{fig:ADULT}
    \vspace{-2ex}
\end{figure}

\revise{We first consider $\omega = \infty$.}
In Fig.~\ref{fig:ADULT}(a), we show the trade-off between the estimated \gls{FNR} and \gls{FPR} for the ADULT dataset, achieved by varying the threshold $\theta$ in~\eqref{eq:test_metric_normal_approx} for $n+1\in \{60, 70, 90\}$ clients. Both the \gls{FNR} and \gls{FPR} can be as small as $0.005$ simultaneously. Hence, the server can reliably distinguish the selected input pair, and the membership inference attack is successful. 
We note that a reference for the trade-off curve that represents different privacy levels is given in~\cite[Fig.~3]{Don21}. There, the case with both \gls{FNR} and \gls{FPR} equal to $0.07$ is already considered nonprivate. Comparing Fig.~\ref{fig:ADULT}(a) with this reference, we conclude that \gls{SecAgg} provides essentially no privacy for the ADULT dataset. 
Next, in Fig.~\ref{fig:ADULT}(b), we show the average audited \gls{LDP} curves 
for the ADULT dataset. We observe that the audited \gls{LDP} curves are close to the largest auditable $(\overline \epsilon_\delta,\delta)$ with the considered $N_{\rm s}$ and~$\gamma$. As $\epsilon$ increases, $\delta$ remains high until it drops due to the limit of the Clopper-Pearson method\textemdash even for $\epsilon=7$, $\delta>10^{-1}$. Furthermore, increasing $n$ provides only a marginal privacy improvement. This shows that the privacy of \gls{SecAgg}, viewed through the lens of \gls{LDP}, is weak.

\begin{figure}[t!]
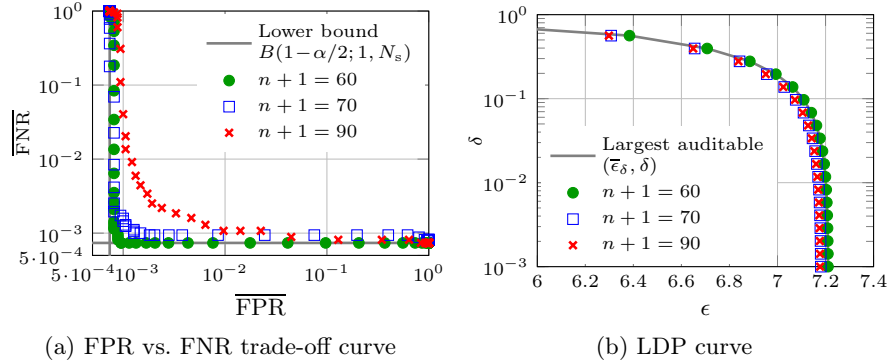

    \centering
    \include{fig/tradeoff_fedavg_ideal}
    \vspace{-4ex}
\caption{Audited \gls{FPR} vs. \gls{FNR} trade-off and \gls{LDP} curves, averaged over $5$ initial models, for  federated learning with \gls{SecAgg} on the EMNIST Digits dataset \revise{with homogeneous data partitioning}. Here, $d = 7850$.}
    \label{fig:EMNIST}
    \vspace{-3ex}
\end{figure}

In Fig.~\ref{fig:EMNIST}, we show the \gls{FPR} vs. \gls{FNR} trade-off and the audited \gls{LDP} curve for the EMNIST Digits dataset. Similar conclusions hold: \gls{SecAgg} provides weak privacy. In this case, with a larger model size than the ADULT dataset, the adversary achieves even smaller \gls{FPR} and \gls{FNR} simultaneously.


\subsubsection{Heterogeneous data partitioning.}
\begin{figure}[t!]
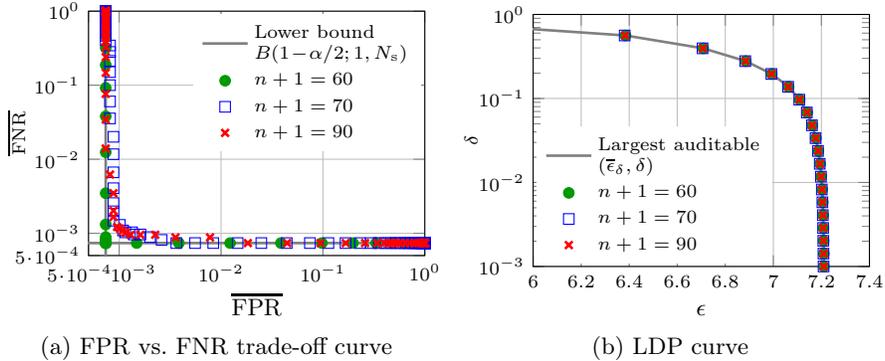

    \centering
    \include{fig/fig_EMNIST_hetero}
    \vspace{-4ex}
\caption{\revise{Same as Fig.~\ref{fig:EMNIST} but with heterogeneous data partitioning.}}
    \label{fig:EMNIST_hetero}
    \vspace{-3ex}
\end{figure}

\revise{We next consider $\omega = 1$ and show the \gls{FPR} vs. \gls{FNR} trade-off and the audited \gls{LDP} curve for the EMNIST Digits dataset in Fig.~\ref{fig:EMNIST_hetero}. In this case, the \gls{FPR} and \gls{FNR} are simultaneously reduced with respect to the homogeneous case, and the audited $(\epsilon,\delta)$ coincide with the largest auditable values. This is because the worst-case pair $(\vx_0,\vx_0')$ is better separated than in the homogeneous case and thus easier to distinguish.}

\vspace{1ex}

\subsubsection{Discussion.} We have seen that \gls{SecAgg} is expected to perform like a correlated Gaussian mechanism (see Section~\ref{sec:asymptotic}). Why does \gls{SecAgg} fail to prevent membership inference attacks, given that the Gaussian mechanism (with appropriate noise calibration) is known to be effective? We explain it as follows. We assume that the individual updates have entries with a bounded magnitude such that 
$\|\rvx_i\|_2 \le r \sqrt{d}$, $i\in [0:n]$. For large $n$, we expect the mechanism $M$ in~\eqref{eq:mechanism_client_0} to have similar privacy guarantee as $G$ in Theorem~\ref{th:Gaussian_correlated_noise} with 
$\sS_d = \{\vx \in \bR^d\colon \|\vx\|_2 \le r \sqrt{d}\}$ and $\mSigma_
\rvy$ being the covariance matrix of $\sum_{i=1}^n \rvx_i$. 
In this case, $\Delta 
\ge 2\sqrt{d/n}$ 
(see Appendix~\ref{proof:Gaussian_correlated_noise_l2}). 
A strong privacy guarantee requires~$\Delta$ to be small, which implies that~$d/n$ must be small. This suggests that the privacy guarantee of \gls{SecAgg} is weak if the vector dimension~$d$ is large compared to the number of clients~$n$. 
This is, however, the case in most practical \gls{FL} scenarios, 
as the model size is typically much larger than the number of clients.  Note that reducing the ratio $d/n$ by dividing the model into smaller chunks that are federated via \gls{SecAgg} does not solve this issue, as the privacy loss composes over these chunks. 

\revise{While with our results we have shown that \gls{SecAgg} provides weak privacy for small models where $d$ is in the order of $10^2$--$10^3$, we remark that the privacy guarantee is expected to further deteriorate for larger models. This is supported by the rapid degradation of the privacy guarantee with $d$ of the Gaussian mechanism (see Fig.~\ref{fig:priv_curve}) and the similarity of SecAgg and a Gaussian mechanism.}

\section{Conclusions} \label{sec:discussion}

We analyzed the privacy of \gls{SecAgg} through the lens of \gls{LDP}. Via privacy auditing, we showed that membership inference attacks on the output of \gls{SecAgg} succeed with high probability: adding independent local updates is not sufficient to hide a local update when the model is of high dimension. While this result may not be surprising, our work fills an important gap by providing a formal analysis of the privacy of \gls{SecAgg} and challenges the prevailing claims of the privacy robustness of \gls{SecAgg}. Hence, it underscores that additional privacy mechanisms, such as noise addition, are needed in federated learning.


\begin{credits}
\subsubsection{\ackname} 
\revise{This work has received funding from the European Union’s Horizon 2020 research and innovation programme under the Marie Sklodowska-Curie grant agreement No 101022113. This work was also partially supported by the 
Wallenberg AI, Autonomous Systems and Software Program
(WASP) and by the Swedish Research Council (VR) under grant 2020-03687. The authors would like to thank Onur Günlü and Balázs Pejó  for fruitful discussions.} 

\end{credits}


\begin{appendices}
\renewcommand{\thesection}{\appendixname~\Alph{section}}
\renewcommand{\thesubsection}{\Alph{section}.\arabic{subsection}}



\section{Correlated Gaussian Mechanism} \label{app:correlated_Gaussian}
We present an analysis of the \gls{LDP} guarantee of the Gaussian mechanism 
    $G(\rvx) = \rvx + \rvy$,  
where $\rvx$ belongs to a subset $\sS_d$ of $\bR^d$, and $\rvy \sim \sN(\mathbf{0}, \mSigma_\rvy)$. 

\subsection{Optimal \gls{LDP} Curve: Proof of Theorem~\ref{th:Gaussian_correlated_noise}} \label{proof:Gaussian_correlated_noise}
    We extend~\cite[Thm.~8]{Bal18} to the case of correlated noise. 
    First, for a mechanism $M$ and a pair $\vx,\vx'$, we define the privacy loss function as $L_{M,\vx,\vx'}(\vz) \triangleq \ln \frac{P_{M(\vx)}(\vz)}{P_{M(\vx')}(\vz)}$. The \gls{PLRV} $\rL_{M,\vx,\vx'}$ is defined as the output of $L_{M,\vx,\vx'}$ when the input 
    follows~$P_{M(\vx)}$. The \gls{PLRV} can be used to express the optimal \gls{LDP} curve as 
        \begin{equation} \label{eq:BalleWang_LDP}
            \delta_M(\epsilon) = \textstyle\max_{\vx,\vx' \in \sS_d} \big(\Prob{\rL_{M,\vx,\vx'} \ge \epsilon} - e^\epsilon \Prob{\rL_{M,\vx',\vx} \le -\epsilon}\big).
        \end{equation}
    Equation~\eqref{eq:BalleWang_LDP} is obtained from a similar result for \gls{DP} given in~\cite[Thm. 5]{Bal18}, upon modifying the notion of neighboring datasets.
    
    For the mechanism $G$, we have that 
    $P_{G(\vx)}(\vz) = \frac{\exp(-\frac{1}{2}(\vz - \vx)^\T \mSigma_\rvy^{-1} (\vz-\vx))}{\sqrt{(2\pi)^d|\mSigma_\rvy|}}$.
       Therefore, the \gls{PLRV} can be expressed as
    \begin{align}
        \rL_{G,\vx,\vx'} 
        &= \frac{1}{2}(\rvz - \vx')^\T \mSigma_\rvy^{-1} (\rvz-\vx') -\frac{1}{2}(\rvz - \vx)^\T \mSigma_\rvy^{-1} (\rvz -\vx) \\
        &= \frac{1}{2}(\vx - \vx')^\T \mSigma_\rvy^{-1} (\vx-\vx') + (\vx - \vx')^\T \mSigma_\rvy^{-1} (\rvz - \vx)\,.
    \end{align}
    With $\rvz$ identically distributed to $G(\vx)$, i.e., $\rvz \sim \sN(\vx,\mSigma_\rvy)$, we have that $\rL_{G,\vx,\vx'} \sim \sN(\eta,2\eta)$ with
        $\eta = \frac{1}{2}(\vx - \vx')^\T \mSigma_\rvy^{-1} (\vx-\vx')$.
    It follows from~\cite[Lemma 7]{Bal18} that for $\rx \sim \sN(\eta,2\eta)$, $\Prob{\rx \ge \epsilon} - e^\epsilon \Prob{\rx \le -\epsilon}$ is monotonically increasing in $\eta$. By applying this result with $\rx = \rL_{G,\vx,\vx'}$, we obtain that the maximum in the right-hand side of~\eqref{eq:BalleWang_LDP} is achieved when $\eta$ is maximized, i.e., when $\eta = \max_{\vx,\vx'\in \sS_d} \frac12(\vx - \vx')^\T \mSigma_\rvy^{-1} (\vx-\vx') = \Delta^2/2$. We then obtain the optimal \gls{LDP} curve~\eqref{eq:priv_curve_Gaussian_correlated_noise} after some simple computations. 

\subsection{The Case $\sS_d = \{\vx \in \bR^d\colon \|\vx\|_2 \le r \sqrt{d}\}$} \label{proof:Gaussian_correlated_noise_l2}
    In this case, for every $\vx,\vx'\in\sS_d$, we have that
    \begin{align}
        (\vx - \vx')^\T \mSigma_\rvy^{-1} (\vx-\vx') &= \|\vx-\vx'\|_2^2 \cdot \frac{(\vx - \vx')^\T}{\|\vx-\vx'\|_2} \mSigma_\rvy^{-1} \frac{\vx-\vx'}{\|\vx-\vx'\|_2} \\
        &\le (\|\vx\|_2 + \|\vx'\|_2)^2 \cdot \lambda\sub{max}(\mSigma_\rvy^{-1}) \label{eq:tmp1170}\\
        &\le (2 r\sqrt{d})^2 \lambda\sub{min}^{-1}(\mSigma_\rvy),\label{eq:tmp1171}
    \end{align}
    where $\lambda\sub{max}(\mSigma_\rvy^{-1})$ is the largest eigenvalue of $\mSigma_\rvy^{-1}$ and  $\lambda\sub{min}(\mSigma_\rvy)$ is the smallest eigenvalue of $\mSigma_\rvy$. 
    Here,~\eqref{eq:tmp1170} follows from the triangle inequality and the Rayleigh-Ritz theorem, and~\eqref{eq:tmp1171} holds because both $\|\vx\|_2$ and $\|\vx'\|_2$ are bounded by $r\sqrt{d}$. Equalities occur in~\eqref{eq:tmp1170} and~\eqref{eq:tmp1171} if $\vx = -\vx' = r\sqrt{d} \vv_{\min}$ where $\vv\sub{min}$ is the eigenvector of $\mSigma_\rvy$ corresponding to $\lambda\sub{min}^{-1}(\mSigma_\rvy)$. Therefore, $\Delta = 2r\sqrt{d/\lambda_{\min}(\mSigma_\rvy)}$. 
    
    If we let $\mSigma_\rvy$ equal the covariance matrix of the sum of $n$ independent vectors $\rvx_1, \dots, \rvx_n$ in $\sS_d$, it holds that
        $\lambda\sub{min}(\mSigma_\rvy) \le 
        \frac{1}{d}\Tr(\mSigma_\rvy) = 
        \frac{1}{d}\Exp{\Tr\left(\sum_{i=1}^n \rvx_i \rvx_i^\T\right)} 
        = \frac{1}{d} \sum_{i=1}^n \Exp{\Tr(\rvx_i \rvx_i^\T)} = \frac{1}{d} \sum_{i=1}^n \Exp{\|\rvx_i\|_2^2} \le n r^2.$
    As a consequence, $\Delta \ge 2\sqrt{d/n}$.

\subsection{Trade-Off Function: Proof of Proposition~\ref{prop:tradeoff_Gaussian}} \label{proof:tradeoff_Gaussian}
   The \gls{LLR} for the test between $\{H \colon \vz$  is generated from $\sN(\vx, \mSigma_\rvy)\}$ and $\{H' \colon \vz$  is generated from $\sN(\vx', \mSigma_\rvy)\}$ is given by the privacy loss function $L_{G,\vx,\vx'}(\vz)$. 
   For a threshold $\theta$, the \gls{FNR} and \gls{FPR} are given by 
   \begin{align}
       \FNR(\theta) &= \Prob[\rvz \sim \sN(\vx',\mSigma_\rvy)]{L_{G,\vx,\vx'}(\rvz) \ge \theta} = \Prob{\rL_{G,\vx',\vx} \le -\theta}, \\ 
       \FPR(\theta) &= \Prob[\rvz \sim \sN(\vx,\mSigma_\rvy)]{L_{G,\vx,\vx'}(\rvz) < \theta} = \Prob{\rL_{G,\vx,\vx'} < \theta}.
    \end{align}
    In the proof of Theorem~\ref{th:Gaussian_correlated_noise}, we have shown that $\rL_{G,\vx,\vx'}(\rvz) \sim \sN(\eta,2\eta)$ with $\eta = \frac{1}{2}(\vx - \vx')^\T \mSigma_\rvy^{-1} (\vx-\vx')$. Therefore, $\FNR(\theta) =\Phi(\frac{-\theta - \eta}{\sqrt{2\eta}})$ and $\FPR(\theta) = \Phi(\frac{\theta - \eta}{\sqrt{2\eta}})$.
    To achieve $\FPR(\theta) \le \alpha$, the threshold must satisfy $\theta \le  \eta - \sqrt{2\eta} \Phi^{-1}(1-\alpha)$. Under this constraint, the minimum FNR is given by $\Phi(\Phi^{-1}(1 - \alpha) -\sqrt{2\eta})$. This is by definition the optimal trade-off $T_{\sN(\vx,\mSigma), \sN(\vx',\mSigma)}(\alpha)$. 

\section{\gls{LDP} Analysis of the Mechanism~\eqref{eq:mechanism_client_0} in a Special Case: Proof of Theorem~\ref{th:dominating_pair}} \label{app:dominating_pair}

To prove Theorem~\ref{th:dominating_pair}, we shall use the following preliminary results. 
\begin{lemma}\label{lem:Hockey_stick_tradeoff}
    For two pairs of distributions $(P_1,Q_1)$ and $(P_2,Q_2)$, if $T_{P_1,Q_1}(\alpha) \ge T_{P_2,Q_2}(\alpha)$ for every $\alpha \in [0,1]$, then $\sfE_{e^\epsilon}(P_1\|Q_1) \le \sfE_{e^\epsilon}(P_2\|Q_2)$ for every $\epsilon > 0$.
\end{lemma}
Lemma~\ref{lem:Hockey_stick_tradeoff} follows directly by expressing the hockey-stick divergence in terms of the trade-off function as follows.
\begin{lemma}\label{lem:hockey_stick_property}
    Consider two distributions $P$ and $Q$ defined over $\sX$. Define a random variable $\rL_P = \ln \frac{Q(\rx)}{P(\rx)}$, $\rx \sim P$, and denote its \gls{CDF} by $F_P(x) = \Prob{\rL_P \le x}$. 
    It holds that
        $\sfE_{e^\epsilon}(P\|Q) = F_P(-\epsilon) - e^\epsilon T_{P,Q}(1- F_P(-\epsilon))$.
\end{lemma}
\emph{Proof of Lemma~\ref{lem:hockey_stick_property}.}
    Define the random variable $\rL_Q = \ln \frac{Q(\rx)}{P(\rx)}$, $\rx \sim Q$, and denote its \gls{CDF} by $F_Q(x)$. 
    Observe that $1- F_P(\theta)$ and $F_Q(\theta)$ are the \gls{FPR} and \gls{FNR} of the likelihood test between $P$ and $Q$ with threshold $\theta$. It follows from Definition~\ref{def:trade_off} and the Neyman-Pearson lemma~\cite[Thm.~8.6.1]{Hog15} that 
    \begin{equation}
        T_{P,Q}(1 - F_P(\theta)) = F_Q(\theta). \label{eq:tmp1025}
    \end{equation}
    We further have that
    \begin{align}
        \sfE_{e^\epsilon} (P \|Q) &= \int_{\sX}\big(P(x) - e^\epsilon Q(x)\big)^+ \dif x \notag \\
        &= \int_{\sX}\big(P(x) - e^\epsilon Q(x)\big) \ind{P(x) \ge e^\epsilon Q(x)} \dif x \notag \\
        &= \int_{\sX} P(x) \ind{P(x) \ge e^\epsilon Q(x)} \dif x -  e^\epsilon\int_{\sX}Q(x) \ind{P(x) \ge  e^\epsilon Q(x)} \dif x \notag \\
        &= F_P(-\epsilon) - e^{\epsilon} F_Q(-\epsilon).  \label{eq:tmp1032}
    \end{align}
    By substituting~\eqref{eq:tmp1025} with $\theta = -\epsilon$ into~\eqref{eq:tmp1032}, we complete the proof.
    \qed

We are now ready to prove Theorem~\ref{th:dominating_pair}. For brevity, we omit the subscript~$0$ in $\rvx_0$ and $\rx_{0j}$, $j\in [d]$.
\subsubsection{The Univariate Case.}
We first consider $d = 1$. Then $M$ in~\eqref{eq:mechanism_client_0} is written as 
\begin{equation}
    M(x) = x + \ry \label{eq:discrete_mechanism_univariate}
\end{equation}
where $\underline r \le x \le \overline r$ and $\ry$ follows a symmetric and log-concave distribution $P_\ry$. We next show that a dominating pair of distribution for this mechanism is $(P_{\underline r+\ry},P_{\overline r+\ry})$. By symmetry of $P_\ry$ around $y_0$, we have that, for every $a, b \in \bR$,
\begin{align}
    \sfE_{e^\epsilon}(a + \ry\| b + \ry) &= \sup_{\sA} P_{a + \ry}(\sA) - e^\epsilon P_{b + \ry}(\sA) \\
    &= \sup_{\sA} P_{\ry}(\sA - a) - e^\epsilon P_{\ry}(\sA - b) \\
    &= \sup_{- \sA + z_0} P_{\ry}(- \sA + z_0 - a) - e^\epsilon P_{\ry}(-\sA + z_0 - b) \\
    &= \sup_{\sA} P_{\ry}(\sA + a + z_0) - e^\epsilon P_{\ry}(\sA + b + z_0) \\
    &= \sup_{\sA' = \sA + a + b + z_0} P_{\ry}(\sA' - b) - e^\epsilon P_{\ry}(\sA' - a) \\
    &= \sup_{\sA} P_{\ry}(\sA - b) - e^\epsilon P_{\ry}(\sA - a) \\
    &= \sfE_{e^\epsilon}(b + \ry\| a + \ry).
\end{align}
Therefore, 
it suffices to prove that 
\begin{equation}
    \sfE_{e^\epsilon}(a + \ry \| b + \ry) \le \sfE_{e^\epsilon}(\underline r + \ry \| \overline r + \ry), \quad \forall \underline r \le a \le b \le \overline r. \label{eq:tmp243}
\end{equation}
To this end, we shall show that $\sfE_{e^\epsilon}(a + \ry \| b + \ry)$ increases with $b - a$, and thus maximized when $(a,b) = \displaystyle\argmax_{a', b' \in [\underline r, \overline r], a' \le b'}(b' - a') = (\underline r, \overline r)$. In light of Lemma~\ref{lem:Hockey_stick_tradeoff}, it suffices to show that $T_{a + \ry, b + \ry}(x)$ decreases with $b - a$ for all $x \in \bR$. Indeed, this is true because $T_{a + \ry, b + \ry}(\alpha) = F_\ry(F_\ry^{-1}(1-\alpha) - (b  -  a))$ for $\ry$ following a log-concave distribution 
(this follows from~\cite[Prop.~A.3]{Don21}), 
and because the \gls{CDF} $F_\ry(\cdot)$ is an increasing function. 

\subsubsection{The Multivariate Case.} \label{app:LDP_dominating_pair_multivariate}
We now address the general case with $d\ge 1$. Using the independence assumption, we can write the mechanism~\eqref{eq:mechanism_client_0} as
    $M(\vx) 
    = (M_1(\vx), M_2(\vx), \dots, M_d(\vx))$
where $M_j(\vx) = \ve_j^\T \vx + \ry _j$, with $\ve_j$ being the $j$th $d$-dimensional canonical basis vector. Therefore, $M(\vx)$ is a (nonadaptive) composition of $d$ mechanisms $\{M_j\}_{j\in [d]}$. 
Observe that each mechanism $M_j$ has the form~\eqref{eq:discrete_mechanism_univariate}. Therefore, $(P_{\underline r_j + \ry_j}, P_{\overline r_j + \ry_j})$ is a dominating pair of distributions for~$M_j$. 
The proof is completed by applying~\cite[Thm.~10]{Zhu22}, which states that if $(P_j,Q_j)$ is a dominating pair of distributions for mechanism $M_j$, $j \in [d]$, then $(P_1 \times \dots \times P_d, Q_1 \times \dots \times Q_d)$ is a dominating pair of distributions for mechanism $M(\vx) = (M_1(\vx), \dots, M_d(\vx))$. 

\end{appendices}

\bibliographystyle{splncs04}
%





\end{document}

%% file: math_commands_Hoang.tex
\usepackage{amsmath,amsfonts,bm,bbm}

\def\eqref#1{(\ref{#1})}

\def\1{\bm{1}}

\def\eps{{\epsilon}}
\newcommand{\given}{\,\vert\,}	
\newcommand{\sub}[1]{_{\mathrm{#1}}}

\newcommand{\Exp}[2][]{\mathbb{E}_{#1}\left[#2\right]}
\newcommand{\Prob}[2][]{\mathbb{P}_{#1}\left[#2\right]}

\newcommand{\T}{{\scriptscriptstyle\mathsf{T}}}

\newcommand*\dif{\mathop{}\mathrm{d}}
\newcommand{\ind}[1]{{\mathbbm{1}{\left\{#1\right\}}}}

\def\rx{{\textnormal{x}}}
\def\ry{{\textnormal{y}}}
\def\rz{{\textnormal{z}}}

\def\rL{{\textnormal{L}}}

\def\rvx{{\mathbf{x}}}
\def\rvy{{\mathbf{y}}}
\def\rvz{{\mathbf{z}}}

\def\vmu{{\bm{\mu}}}

\def\ve{{\bm{e}}}

\def\vv{{\bm{v}}}

\def\vx{{\bm{x}}}
\def\vy{{\bm{y}}}
\def\vz{{\bm{z}}}

\def\mI{{\bm{I}}}

\def\mSigma{{\bm{\Sigma}}}

\DeclareMathAlphabet{\mathsfit}{\encodingdefault}{\sfdefault}{m}{sl}
\SetMathAlphabet{\mathsfit}{bold}{\encodingdefault}{\sfdefault}{bx}{n}

\def\bR{{\mathbb{R}}}

\def\sfE{{\mathsf{E}}}

\def\sA{{\mathcal{A}}}

\def\sE{{\mathcal{E}}}

\def\sN{{\mathcal{N}}}

\def\sS{{\mathcal{S}}}

\def\sX{{\mathcal{X}}}

\newcommand{\Cov}{\mathrm{Cov}}

\DeclareMathOperator*{\argmax}{arg\,max}
\DeclareMathOperator*{\argmin}{arg\,min}

\DeclareMathOperator{\Tr}{Tr}

%% file: fig/priv_curve_unif_n20.tex
\subfloat[$\rvx_i$, $i\in {[n]}$, has independent entries exponentially distributed with parameter $1$; $0 \le \rx_{0j} \le 4$, $j\in {[d]}$. For the Gaussian mechanism, $\sS_d = {[0,4]}^d$ and $\mSigma_\rvy = n\mI_d$. 
]{
 \label{fig:exponential}%
\begin{tikzpicture}[scale = .8]
		\begin{axis}[%
			width=2.3in,
			height=1.5in,
			at={(0.759in,0.481in)},
			scale only axis,
			xmin=0,
			xmax=10,
			xlabel={$\epsilon$},
			ymin=1e-3,
			ymax=1,
			yminorticks=true,
            ymode=log,
			ylabel style={font=\color{black}, yshift=-2ex},
			ylabel={$\delta$},
			xmajorgrids,
			ymajorgrids,
            legend style={at={(.01,.01)}, anchor=south west, legend cell align=left, align=left,draw=none, fill=white, fill opacity=.7,text opacity = 1}
			]		
  
            \addplot [line width = 1,color=red,mark = x, mark color = red,only marks,mark size = 3]
		table [x index = {0}, y index={1}, col sep=comma]
		{./fig/priv_curve_exp_n300.csv}; 
        \addlegendentry{Exponential};

                        \addplot [line width = 1,color=blue]
		table [x index = {0}, y index={2}, col sep=comma]
		{./fig/priv_curve_exp_n300.csv}; 
        \addlegendentry{Gaussian mechanism};
  

            \addplot [line width = 1,color=blue,forget plot]
		table [x index = {0}, y index={4}, col sep=comma]
		{./fig/priv_curve_exp_n300.csv}; 
  
            \addplot [line width = 1,color=red,mark = x, mark color = red,only marks,mark size = 3,forget plot]
		table [x index = {0}, y index={3}, col sep=comma]
		{./fig/priv_curve_exp_n300.csv}; 

  
  
            \addplot [line width = 1,color=blue,forget plot]
		table [x index = {0}, y index={6}, col sep=comma]
		{./fig/priv_curve_exp_n300.csv}; 
  
            \addplot [line width = 1,color=red,mark = x, mark color = red,only marks,mark size = 3,forget plot]
		table [x index = {0}, y index={5}, col sep=comma]
		{./fig/priv_curve_exp_n300.csv}; 

  

  

            \node[align = center] at (axis cs:5.7,1.5e-2) {$d = 100$};
            \node[align = center] at (axis cs:7.8,8e-2) {$d = 200$};
            \node[align = center] at (axis cs:8.7,6e-1) {$d = 300$};

            \node[draw] () at (axis cs:2,9e-3) {$n = 300$};
		\end{axis}
	\end{tikzpicture}%
 }
 \hspace{.2cm}
 \subfloat[$\rvx_i$, $i\in {[n]}$, has independent entries uniformly distributed in ${[-1/2,1/2]}$; $-1/2 \le \rx_{0j} \le 1/2$, $j\in {[d]}$. For the Gaussian mechanism, $\sS_d = {[-1/2,1/2]}^d$ and $\mSigma_\rvy = \frac{n}{12}\mI_d$. 
 ]{
 \label{fig:uniform}%
\begin{tikzpicture}[scale = .8]
		\begin{axis}[%
			width=2.3in,
			height=1.5in,
			at={(0.759in,0.481in)},
			scale only axis,
			xmin=0,
			xmax=10,
			xlabel={$\epsilon$},
			ymin=1e-3,
			ymax=1,
			yminorticks=true,
            ymode=log,
			ylabel style={yshift=-2ex},
			ylabel={$\delta$},
			xmajorgrids,
			ymajorgrids,
            legend style={at={(.01,.01)}, anchor=south west, legend cell align=left, align=left,draw=none, fill=white, fill opacity=.7,text opacity = 1}
			]		
  
  
            \addplot [line width = 1,color=red,mark = x, mark color = red,only marks,mark size = 3]
		table [x index = {0}, y index={3}, col sep=comma]
		{./fig/priv_curve_unif_n20.csv}; 
        \addlegendentry{Uniform};
        
                        \addplot [line width = 1,color=blue]
		table [x index = {0}, y index={4}, col sep=comma]
		{./fig/priv_curve_unif_n20.csv}; 
        \addlegendentry{Gaussian mechanism};
        
            \addplot [line width = 1,color=blue,forget plot]
		table [x index = {0}, y index={6}, col sep=comma]
		{./fig/priv_curve_unif_n20.csv}; 
  
            \addplot [line width = 1,color=red,mark = x, mark color = red,only marks,mark size = 3,forget plot]
		table [x index = {0}, y index={5}, col sep=comma]
		{./fig/priv_curve_unif_n20.csv}; 

            \addplot [line width = 1,color=blue,forget plot]
		table [x index = {0}, y index={8}, col sep=comma]
		{./fig/priv_curve_unif_n20.csv}; 
  
            \addplot [line width = 1,color=red,mark = x, mark color = red,only marks,mark size = 3,forget plot]
		table [x index = {0}, y index={7}, col sep=comma]
		{./fig/priv_curve_unif_n20.csv}; 

            \node[align = center] at (axis cs:6.5,1.5e-2) {$d = 10$};
            \node[align = center] at (axis cs:7.8,1.2e-1) {$d = 20$};
            \node[align = center] at (axis cs:9,4.5e-1) {$d = 40$};

            \node[draw] () at (axis cs:2,9e-3) {$n = 20$};
		\end{axis}
	\end{tikzpicture}%
 }

%% file: fig/fig_ADULT.tex
\subfloat[FPR vs. FNR trade-off curve \label{fig:tradeoff_EMNIST}]{
\begin{tikzpicture}[]
		\tikzstyle{every node}=[font=\scriptsize]
		\begin{axis}[%
            scale = .8,
			width=2.2in,
			height=1.6in,
			at={(0.759in,0.481in)},
			scale only axis,
			xmin=5e-4,
			xmax=1,
			xtick={1,1e-1, 1e-2, 1e-3, 5e-4},
            xticklabels={$10^0$,$10^{-1}$, $10^{-2}$, $\quad ~10^{-3}$, $\!\!\!\!\!5 \!\cdot\! 10^{-4}$},
			xlabel style={font=\color{black},yshift=1ex},
			xlabel={$\overline \FPR$},
			ymin=5e-4,
			ymax=1,
            xmode=log,
            ymode=log,
			ytick={1,1e-1, 1e-2, 1e-3, 5e-4},
            yticklabels={$10^0$,$10^{-1}$, $10^{-2}$, $10^{-3}$, $5 \!\cdot\! 10^{-4}$},
			ylabel style={font=\color{black}, yshift=-2ex},
			ylabel={\scriptsize$\overline \FNR$},
			xmajorgrids,
			ymajorgrids,
            legend style={at={(.99,.99)}, anchor=north east, legend cell align=left, align=left,draw=none, fill=white, fill opacity=.6,text opacity = 1}
			]		

            \addplot [line width = 1, color=gray]
			table[row sep=crcr]{%
				1e-4 0.000737503801108105\\
                1 0.000737503801108105 \\
			};

            \addplot [line width = 1, color=gray,forget plot]
			table[row sep=crcr]{%
				0.000737503801108105 1e-4 \\
                0.000737503801108105 1\\
			};
            \addlegendentry{Lower bound \\ $B(1\!-\!\alpha/2; 1, N_{\rm s})$};

            \addplot [mark=*,color=OliveGreen,only marks,mark size=2,mark repeat={6}]
		      table [x index = {4}, y index={1}, col sep=comma]
		      {./fig/tradeoff_SecAgg_ADULT_n-60_alpha-inf.csv}; 
              \addlegendentry{$n + 1 = 60$};

            \addplot [mark=square,color=blue,only marks,mark size=2,mark repeat={6}]
		      table [x index = {4}, y index={1}, col sep=comma]
		      {./fig/tradeoff_SecAgg_ADULT_n-70_alpha-inf.csv}; 
              \addlegendentry{$n + 1 = 70$};

              \addplot [line width = 1,color=red,only marks,mark size=2, mark = x,mark options=solid,mark repeat={6}]
		      table [x index = {4}, y index={1}, col sep=comma]
		      {./fig/tradeoff_SecAgg_ADULT_n-90_alpha-inf.csv}; 
              \addlegendentry{$n + 1 = 90$};
		\end{axis}
	\end{tikzpicture}%
 }
 \subfloat[\gls{LDP} curve]{
 \label{fig:LDP_EMNIST}%
 \begin{tikzpicture}[]
		\tikzstyle{every node}=[font=\scriptsize]
		\begin{axis}[%
            scale = .8,
			width=2.2in,
			height=1.65in,
			at={(0.759in,0.481in)},
			scale only axis,
			xmin=6,
			xmax=7.4,
			xtick={6, 6.2, 6.4, 6.6, 6.8, 7, 7.2, 7.4},
			xlabel style={font=\color{black},yshift=1ex},
			xlabel={$\epsilon$},
			ymin=1e-3,
			ymax=1,
			yminorticks=true,
            ymode=log,
			ylabel style={font=\color{black}, yshift=-3ex},
			ylabel={\scriptsize $\delta$},
			xmajorgrids,
			ymajorgrids,
            legend style={at={(.01,.01)}, anchor=south west, legend cell align=left, align=left,draw=none, fill=white, fill opacity=.6,text opacity = 1}
			]		

            \addplot [line width=1, color=gray]
		      table [x index = {0}, y index={4}, col sep=comma]
		      {./fig/priv_curve_SecAgg_ADULT_n-60_alpha-inf.csv}; 
              \addlegendentry{Largest auditable \\ $(\overline \epsilon_\delta,\delta)$};

            \addplot [mark=*,color=OliveGreen,only marks,mark size=2]
		      table [x index = {2}, y index={4}, col sep=comma]
		      {./fig/priv_curve_SecAgg_ADULT_n-60_alpha-inf.csv}; 
            \addlegendentry{$n + 1 = 60$};

            \addplot [mark=square,color=blue,only marks,mark size=2]
		      table [x index = {2}, y index={4}, col sep=comma]
		      {./fig/priv_curve_SecAgg_ADULT_n-70_alpha-inf.csv}; 
            \addlegendentry{$n + 1 = 70$};
            
            \addplot [line width = 1,color=red,only marks,mark size=2, mark = x,mark options=solid]
		      table [x index = {2}, y index={4}, col sep=comma]
		      {./fig/priv_curve_SecAgg_ADULT_n-90_alpha-inf.csv}; 
            \addlegendentry{$n + 1 = 90$};
		\end{axis}
	\end{tikzpicture}%
 }

%% file: fig/tradeoff_fedavg_ideal.tex
\subfloat[FPR vs. FNR trade-off curve \label{fig:tradeoff_EMNIST}]{
\begin{tikzpicture}[]
		\tikzstyle{every node}=[font=\scriptsize]
		\begin{axis}[%
            scale = .8,
			width=2.2in,
			height=1.6in,
			at={(0.759in,0.481in)},
			scale only axis,
			xmin=5e-4,
			xmax=1,
			xtick={1,1e-1, 1e-2, 1e-3, 5e-4},
            xticklabels={$10^0$,$10^{-1}$, $10^{-2}$, $\quad ~10^{-3}$, $\!\!\!\!\!5 \!\cdot\! 10^{-4}$},
			xlabel style={font=\color{black},yshift=1ex},
			xlabel={$\overline \FPR$},
			ymin=5e-4,
			ymax=1,
            xmode=log,
            ymode=log,
			ytick={1,1e-1, 1e-2, 1e-3, 5e-4},
            yticklabels={$10^0$,$10^{-1}$, $10^{-2}$, $10^{-3}$, $5 \!\cdot\! 10^{-4}$},
			ylabel style={font=\color{black}, yshift=-2ex},
			ylabel={\scriptsize $\overline \FNR$},
			xmajorgrids,
			ymajorgrids,
            legend style={at={(.99,.99)}, anchor=north east, legend cell align=left, align=left,draw=none, fill=white, fill opacity=.6,text opacity = 1}
			]		

            \addplot [line width = 1, color=gray]
			table[row sep=crcr]{%
				1e-4 0.000737503801108105\\
                1 0.000737503801108105 \\
			};

            \addplot [line width = 1, color=gray,forget plot]
			table[row sep=crcr]{%
				0.000737503801108105 1e-4 \\
                0.000737503801108105 1\\
			};
            \addlegendentry{Lower bound \\ $B(1\!-\!\alpha/2; 1, N_{\rm s})$};

            \addplot [mark=*,color=OliveGreen,only marks,mark size=2,mark repeat={8}]
		      table [x index = {4}, y index={1}, col sep=comma]
		      {./fig/tradeoff_SecAgg_EMNIST_n-60_alpha-inf.csv}; 
              \addlegendentry{$n + 1 = 60$};

            \addplot [mark=square,color=blue,only marks,mark size=2,mark repeat={8}]
		      table [x index = {4}, y index={1}, col sep=comma]
		      {./fig/tradeoff_SecAgg_EMNIST_n-70_alpha-inf.csv}; 
              \addlegendentry{$n + 1 = 70$};

              \addplot [line width = 1,color=red,only marks,mark size=2, mark = x,mark options=solid,mark repeat={8}]
		      table [x index = {4}, y index={1}, col sep=comma]
		      {./fig/tradeoff_SecAgg_EMNIST_n-90_alpha-inf.csv}; 
              \addlegendentry{$n + 1 = 90$};
		\end{axis}
	\end{tikzpicture}%
 }
 \subfloat[\gls{LDP} curve]{
 \label{fig:LDP_EMNIST}%
 \begin{tikzpicture}[]
		\tikzstyle{every node}=[font=\scriptsize]
		\begin{axis}[%
            scale = .8,
			width=2.2in,
			height=1.65in,
			at={(0.759in,0.481in)},
			scale only axis,
			xmin=6,
			xmax=7.4,
			xtick={6, 6.2, 6.4, 6.6, 6.8, 7, 7.2, 7.4},
			xlabel style={font=\color{black},yshift=1ex},
			xlabel={$\epsilon$},
			ymin=1e-3,
			ymax=1,
			yminorticks=true,
            ymode=log,
			ylabel style={font=\color{black}, yshift=-3ex},
			ylabel={\scriptsize $\delta$},
			xmajorgrids,
			ymajorgrids,
            legend style={at={(.01,.01)}, anchor=south west, legend cell align=left, align=left,draw=none, fill=white, fill opacity=.6,text opacity = 1}
			]		

            \addplot [line width=1, color=gray]
		      table [x index = {0}, y index={4}, col sep=comma]
		      {./fig/priv_curve_SecAgg_EMNIST_n-60_alpha-inf.csv}; 
              \addlegendentry{Largest auditable \\ $(\overline \epsilon_\delta,\delta)$};

            \addplot [mark=*,color=OliveGreen,only marks,mark size=2]
		      table [x index = {2}, y index={4}, col sep=comma]
		      {./fig/priv_curve_SecAgg_EMNIST_n-60_alpha-inf.csv}; 
            \addlegendentry{$n + 1 = 60$};

            \addplot [mark=square,color=blue,only marks,mark size=2]
		      table [x index = {2}, y index={4}, col sep=comma]
		      {./fig/priv_curve_SecAgg_EMNIST_n-70_alpha-inf.csv}; 
            \addlegendentry{$n + 1 = 70$};
            
            \addplot [line width = 1,color=red,only marks,mark size=2, mark = x,mark options=solid]
		      table [x index = {2}, y index={4}, col sep=comma]
		      {./fig/priv_curve_SecAgg_EMNIST_n-90_alpha-inf.csv}; 
            \addlegendentry{$n + 1 = 90$};
		\end{axis}
	\end{tikzpicture}%
 }

%% file: fig/fig_EMNIST_hetero.tex
\subfloat[FPR vs. FNR trade-off curve \label{fig:tradeoff_EMNIST}]{
\begin{tikzpicture}[]
		\tikzstyle{every node}=[font=\scriptsize]
		\begin{axis}[%
            scale = .8,
			width=2.2in,
			height=1.6in,
			at={(0.759in,0.481in)},
			scale only axis,
			xmin=5e-4,
			xmax=1,
			xtick={1,1e-1, 1e-2, 1e-3, 5e-4},
            xticklabels={$10^0$,$10^{-1}$, $10^{-2}$, $\quad ~10^{-3}$, $\!\!\!\!\!5 \!\cdot\! 10^{-4}$},
			xlabel style={font=\color{black},yshift=1ex},
			xlabel={$\overline \FPR$},
			ymin=5e-4,
			ymax=1,
            xmode=log,
            ymode=log,
			ytick={1,1e-1, 1e-2, 1e-3, 5e-4},
            yticklabels={$10^0$,$10^{-1}$, $10^{-2}$, $10^{-3}$, $5 \!\cdot\! 10^{-4}$},
			ylabel style={font=\color{black}, yshift=-2ex},
			ylabel={\scriptsize $\overline \FNR$},
			xmajorgrids,
			ymajorgrids,
            legend style={at={(.99,.99)}, anchor=north east, legend cell align=left, align=left,draw=none, fill=white, fill opacity=.6,text opacity = 1}
			]		

            \addplot [line width = 1, color=gray]
			table[row sep=crcr]{%
				1e-4 0.000737503801108105\\
                1 0.000737503801108105 \\
			};

            \addplot [line width = 1, color=gray,forget plot]
			table[row sep=crcr]{%
				0.000737503801108105 1e-4 \\
                0.000737503801108105 1\\
			};
            \addlegendentry{Lower bound \\ $B(1\!-\!\alpha/2; 1, N_{\rm s})$};
              
            \addplot [mark=*,color=OliveGreen,only marks,mark size=2,mark repeat={8}]
		      table [x index = {4}, y index={1}, col sep=comma]
		      {./fig/tradeoff_SecAgg_EMNIST_n-60_alpha-1.csv}; 
              \addlegendentry{$n + 1 = 60$};

            \addplot [mark=square,color=blue,only marks,mark size=2,mark repeat={8}]
		      table [x index = {4}, y index={1}, col sep=comma]
		      {./fig/tradeoff_SecAgg_EMNIST_n-70_alpha-1.csv}; 
              \addlegendentry{$n + 1 = 70$};

              \addplot [line width = 1,color=red,only marks,mark size=2, mark = x,mark options=solid,mark repeat={8}]
		      table [x index = {4}, y index={1}, col sep=comma]
		      {./fig/tradeoff_SecAgg_EMNIST_n-90_alpha-1.csv}; 
              \addlegendentry{$n + 1 = 90$};
		\end{axis}
	\end{tikzpicture}%
 }
 \subfloat[\gls{LDP} curve]{
 \label{fig:LDP_EMNIST}%
 \begin{tikzpicture}[]
		\tikzstyle{every node}=[font=\scriptsize]
		\begin{axis}[%
            scale = .8,
			width=2.2in,
			height=1.65in,
			at={(0.759in,0.481in)},
			scale only axis,
			xmin=6,
			xmax=7.4,
			xtick={6, 6.2, 6.4, 6.6, 6.8, 7, 7.2, 7.4},
			xlabel style={font=\color{black},yshift=1ex},
			xlabel={$\epsilon$},
			ymin=1e-3,
			ymax=1,
			yminorticks=true,
            ymode=log,
			ylabel style={font=\color{black}, yshift=-3ex},
			ylabel={\scriptsize $\delta$},
			xmajorgrids,
			ymajorgrids,
            legend style={at={(.01,.01)}, anchor=south west, legend cell align=left, align=left,draw=none, fill=white, fill opacity=.6,text opacity = 1}
			]		

            \addplot [line width=1, color=gray]
		      table [x index = {0}, y index={4}, col sep=comma]
		      {./fig/priv_curve_SecAgg_EMNIST_n-60_alpha-1.csv}; 
              \addlegendentry{Largest auditable \\ $(\overline \epsilon_\delta,\delta)$};

            \addplot [mark=*,color=OliveGreen,only marks,mark size=2]
		      table [x index = {2}, y index={4}, col sep=comma]
		      {./fig/priv_curve_SecAgg_EMNIST_n-60_alpha-1.csv}; 
            \addlegendentry{$n + 1 = 60$};

            \addplot [mark=square,color=blue,only marks,mark size=2]
		      table [x index = {2}, y index={4}, col sep=comma]
		      {./fig/priv_curve_SecAgg_EMNIST_n-70_alpha-1.csv}; 
            \addlegendentry{$n + 1 = 70$};
            
            \addplot [line width = 1,color=red,only marks,mark size=2, mark = x,mark options=solid]
		      table [x index = {2}, y index={4}, col sep=comma]
		      {./fig/priv_curve_SecAgg_EMNIST_n-90_alpha-1.csv}; 
            \addlegendentry{$n + 1 = 90$};
		\end{axis}
	\end{tikzpicture}%
 }